\documentclass[lettersize,journal]{IEEEtran}
\usepackage{amsmath,amsfonts}
\usepackage{algorithmic}
\usepackage{algorithm}
\usepackage{array}
\usepackage[caption=false,font=normalsize,labelfont=sf,textfont=sf]{subfig}
\usepackage{textcomp}
\usepackage{stfloats}
\usepackage{url}
\usepackage{verbatim}
\usepackage{graphicx}
\usepackage{cite}

\usepackage{algorithm}
\usepackage{algorithmic}
\usepackage{amsmath}
\usepackage{multirow}
\usepackage{amssymb} 
\usepackage{pifont}
\usepackage{float}
\usepackage{placeins}
\usepackage{tabularx,booktabs}
\newcolumntype{Y}{>{\centering\arraybackslash}X}
\usepackage{xcolor}
\begin{document}

\title{Blind Quality Enhancement of Compressed Video via Fine-Grained Degradation-Guided Sequential Inference}

\author{Li Yu,~\IEEEmembership{Member,~IEEE}, Yingbo Zhao, Shiyu Wu, Siyue Yu,~\IEEEmembership{Member,~IEEE}, Moncef Gabbouj,~\IEEEmembership{Fellow,~IEEE}, and Qingshan Liu,~\IEEEmembership{Senior~Member,~IEEE}  
\thanks{This work was supported in part by the National Natural Science Foundation of China under Grant 62002172.}
\thanks{Li Yu is with School of Computer Science, Nanjing University of Information Science \& Technology, Nanjing, China, and also with Jiangsu Collaborative Innovation Center of Atmospheric Environment and Equipment Technology (CICAEET), Nanjing University of Information Science \& Technology, Nanjing, China.}
\thanks{Yingbo Zhao is with School of Computer Science, Nanjing University of Information Science \& Technology, Nanjing, China.}
\thanks{Shiyu Wu is with School of Software, Nanjing University of Information Science \& Technology, Nanjing, China.}
\thanks{Siyue Yu is with department of Intelligent Science, Xi’an Jiaotong-Liverpool University, Suzhou, China}
\thanks{Moncef Gabbouj is with the Department of Computing Sciences, Tampere
University, 33100 Tampere, Finland (e-mail: moncef.gabbouj@tuni.fi).}
\thanks{Qingshan Liu is with School of Computer Science, Nanjing University of Posts and Telecommunications, Nanjing, China.}
\thanks{Corresponding author: Moncef Gabbouj}}

\IEEEpubid{\scriptsize\begin{tabular}{c}
Copyright~\copyright~2026 IEEE. Personal use of this material is permitted.\\ However, permission to use this material for any other purposes must be obtained from the IEEE by sending an email to pubs-permissions@ieee.org.
\end{tabular}}



\maketitle

\begin{abstract}
Existing studies on quality enhancement for compressed video (QECV) predominantly rely on known quantization parameters (QPs), training separate enhancement models for each QP setting, which are referred to as non-blind methods. However, in practical scenarios such as transcoding and transmission, QPs may be partially or entirely unavailable, which limits the applicability of these methods and motivates the development of blind QECV techniques. Existing blind methods typically generate degradation vectors using classification models trained with cross-entropy loss, and employ them as channel attention to guide artifact reduction. Nevertheless, such degradation representations mainly capture global compression information and lack fine-grained spatial cues, making them less effective in handling spatially varying artifact patterns. To address this issue, we propose a pre-trained degradation representation learning module that decouples and extracts high-dimensional, multi-scale degradation representations from compressed video content, providing fine-grained guidance for artifact reduction. Furthermore, most existing blind and non-blind methods adopt a uniform inference architecture for all compression levels, ignoring the distinct computational demands of different QPs. To overcome this limitation, we introduce a sequential inference strategy that adaptively adjusts the number of artifact reduction stages according to the estimated compression level. Extensive experiments show that the proposed method significantly improves enhancement performance. In particular, at QP = 22, it raises PSNR improvement from 0.31 dB to 0.65 dB over the previous state-of-the-art blind method. Meanwhile, with the proposed sequential inference strategy, the average inference time at QP = 22 is reduced by 50\% compared with that at QP = 42.
\end{abstract}

\section{Introduction}
The growing demand for 4K/8K video content faces significant challenges due to bandwidth and storage limitations. To mitigate these constraints, higher compression ratios are commonly employed~\cite{wiegand2003overview,sullivan2012overview,bross2021overview}, often introducing visual artifacts such as blurring, blocking, and ringing effects~\cite{seshadrinathan2010study}. These distortions considerably impair visual quality, underscoring the importance of effective Quality Enhancement for Compressed Video (QECV) in applications like video streaming, surveillance, and online education~\cite{streaming,surveillance}.

\IEEEpubidadjcol

\begin{figure}[!t]
\centerline{\includegraphics[width=1.05\linewidth]{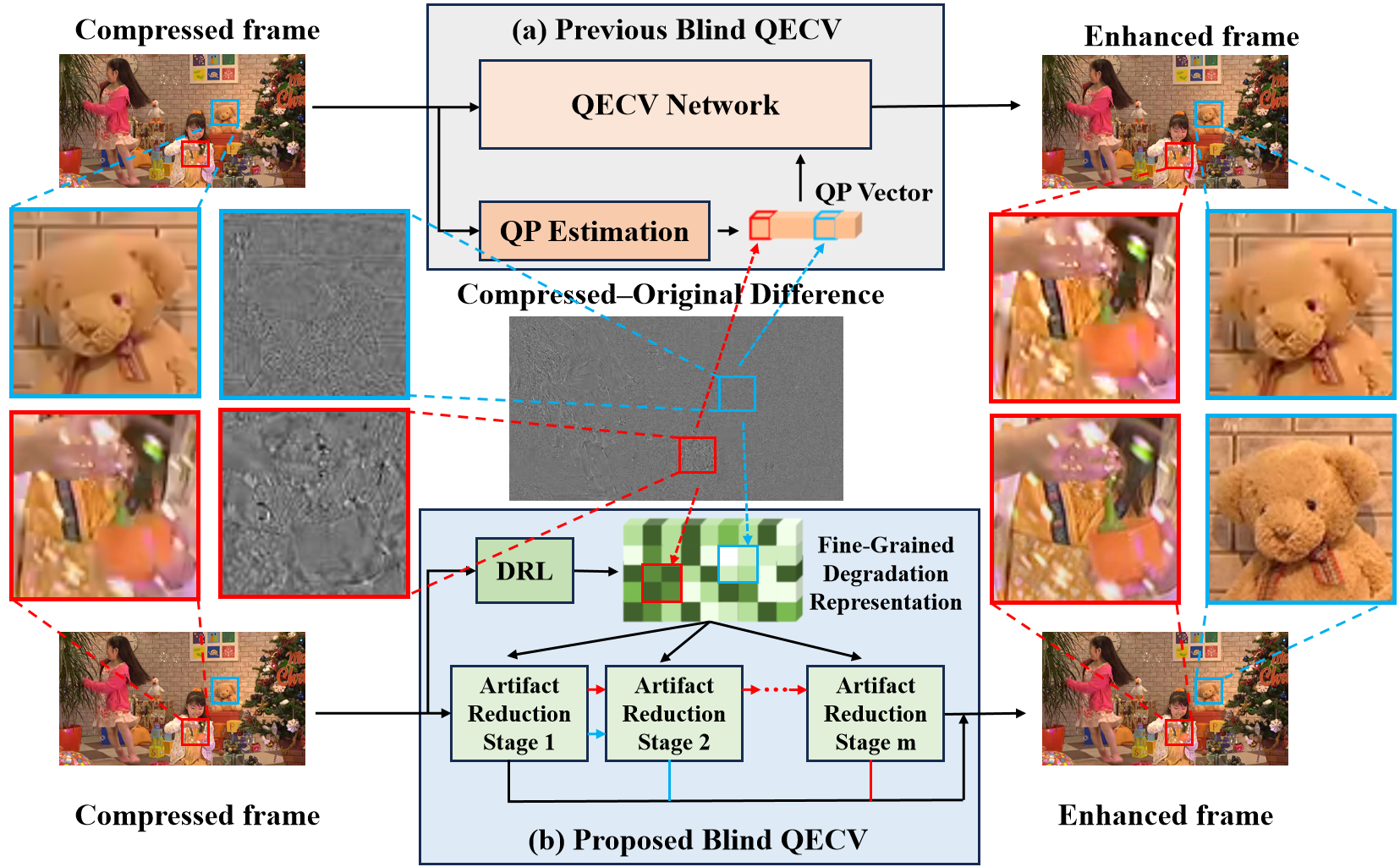}}
\caption{Overview of blind QECV methods. (a) shows existing method that estimates a QP vector to guide artifact removal. (b) presents our method, which extracts fine-grained degradation representation and employs a sequential inference strategy to adaptively perform multi-stage artifact reduction. The difference map in the center highlights spatially varying degradation, where the red region indicates more severe artifacts than the blue one. Our method achieves superior enhancement results on both regions over existing method.}
\label{fig:Introduction}
\end{figure}

While deep learning~\cite{liu2020dlvc} has been widely applied to compressed video enhancement, most existing methods~\cite{Hou_Zhao_Wang_2021,guan2019mfqe,meng2019enhancing,deng2020spatio,zhao2021recursive,xu2022transcoded,yang2019qecnn,dai2017vrcnn,huang2023fastcnn,lin2020asn,hou2017qevideo,li2019mif,zeng2025pnpvcve,he2025rivuletmlp,chen2023fastmfqe,yang2023pimnet,wang2024reconstructionflow,wang2024pixrevive,li2024ctve,meng2020robust,dong2024temporal} adopt non-blind strategies that require training separate models for different quantization parameter (QP) values. This approach increases deployment costs and limits adaptability, particularly in real-world scenarios where QP may be unavailable, such as Digital Rights Management (DRM) that prohibit the retrieval of encoding parameters including QP, post processing and streaming protocols that discard QPs. In such cases, selecting the appropriate enhancement model becomes challenging, driving the need for blind QECV approaches that utilize a single model across all QP settings. Although several blind enhancement methods exist for images~\cite{chen2022blind,jiang2021towards,liu2020liu4k}, they are limited in video applications due to insufficient modeling of spatiotemporal structures and temporal dependencies, making them suboptimal for video enhancement tasks.

For blind video enhancement, existing method \cite{ding2023blind} trains a QP classifier using cross-entropy loss and employs the resulting vector as channel attention to guide artifact removal, as illustrated in Fig. \ref{fig:Introduction}(a). However, degradation patterns often exhibit significant spatial variation within a single frame. For example, in Fig. \ref{fig:Introduction}, the bubble region (red box) is considerably more distorted than the bear region (blue box). A single vector cannot effectively  capture such intricate spatial discrepancies, lacking the granularity and discriminability needed to model complex and spatiotemporally non-uniform compression artifacts. Moreover, current QECV methods generally apply a uniform enhancement process across all input videos, irrespective of compression severity. This results in computational overallocation for lightly compressed videos and inadequate enhancement for heavily compressed ones.

To address these challenges, we propose a fine-grained degradation-guided sequential inference framework for QECV, as illustrated in Fig. \ref{fig:Introduction}(b). At its core, a Degradation Representation Learning (DRL) module is introduced to capture multi-scale, fine-grained spatial variations of compression artifacts. To reduce content–degradation correlation and improve degradation discriminability, we employ a dual-supervision strategy: contrastive learning enhances the discrimination of local distortion patterns by pulling similar artifact regions closer while pushing dissimilar ones apart, while classification learning imposes semantic constraints to stabilize the representation of distortion levels. This joint approach mitigates reliance on large-scale labeled data and strengthens generalization in degradation modeling, effectively addressing the dynamic content-distortion entanglement inherent in video frames.
Leveraging the fine-grained degradation representation predicted by the DRL module, we further design a sequential inference strategy that dynamically allocates computational resources by adjusting the number of artifact reduction stages. As shown in Fig. \ref{fig:Introduction}(b), lightly degraded regions (e.g., the blue box) stop at earlier stages in the sequential inference process, while heavily distorted regions (e.g., the red box) undergo more processing stages, achieving an adaptive balance between performance and efficiency. Each artifact reduction stage leverages fine-grained degradation-guided feature modulation (from DRL) for robust adaptation, followed by a dual-branch architecture that explicitly models both global and local spatio-temporal dependencies using a Transformer and multi-scale dilated convolutions, respectively. The proposed method achieves superior enhancement performance as a result.

In summary, our main contributions are as follows:
\begin{enumerate}
\item A Fine-Grained Degradation-Guided Sequential Inference method for blind QECV is proposed, which dynamically adjusts the artifact reduction stages based on the severity of compression to balance performance and efficiency.

\item A Degradation Representation Learning module is proposed to evaluate the severity of compression, which combines contrastive learning and classification to effectively reduce content–degradation correlation and improve degradation discriminability.

\item For each artifact reduction stage, it first aggregates spatiotemporal features with fine-grained degradation representations from DRL module for degradation-guided feature modulation; and then performs dual-branch fusion integrating global context with local detail to further exploit spatiotemporal dependencies.

\item Our method achieves a PSNR improvement of 0.65 dB at QP = 22, outperforming the previous SOTA blind method by 0.34 dB, and reduces inference time by \textbf{50\%} compared to QP = 42.

\end{enumerate}

\section{Related Work}

\noindent
\textbf{Non-Blind Video Quality Enhancement.}
Recent multi-frame deep learning methods for QECV mainly fall into two types: local motion compensation and global context modeling.
The former aligns neighboring frames using optical flow, motion vectors, or offset fields to aggregate temporal information~\cite{yang2018mfqe,luo2022coarse,peng2022ovqe}, but struggles with long-range dependencies due to limited receptive fields. To mitigate this, deformable convolutions~\cite{deng2020spatio} are introduced to adaptively enhance temporal aggregation. Liu et al.~\cite{liu2024enlarged} proposed an enlarged motion-aware and frequency-aware network that enhanced motion compensation by incorporating enlarged receptive fields and frequency-domain information. The latter leverages frame-wise global similarity. Xu et al.~\cite{xu2019non} proposed a non-local LSTM-based approach, albeit with high computational cost. With the advent of vision Transformers, TVQE~\cite{10332936} applied Swin Transformer to QECV for efficient global modeling. Subsequent works~\cite{zhang2023video,yu2024multi} further improved performance via local-global fusion and advanced window strategies. However, these non-blind methods are highly QP-sensitive and often degrade under mismatched compression levels.

\noindent
\textbf{Blind Quality Enhancement.}
Blind quality enhancement methods~\cite{jiang2021towards,wang2022jpeg,chen2022blind,ding2023blind} aim to address all levels of compression quality using a single model. Most of these methods are designed for image enhancement. ~\cite{wang2022jpeg} achieves accurate removal of JPEG compression artifacts by mining discriminative features of the artifacts through contrastive representation learning, without relying on known JPEG compression parameters. ~\cite{jiang2021towards} instead explicitly decouple image features and QF features from compressed images, and map the predicted QF to modulation coefficients that control the responses of the reconstruction network, thereby enabling flexible blind JPEG artifact removal across different compression levels under unknown QFs. ~\cite{ding2023blind} achieves blind CVQE for the first time by predicting the quality of target frames and assigning different computational weights to the outputs of various compression artifact reduction blocks. However, although QPs are embedded in compressed video bitstreams, they may be unavailable or inaccessible to downstream enhancement models in real-world applications. Therefore, blind CVQE remains practically important.

\noindent
\textbf{Adaptive Computation.}
Existing methods can be roughly divided into two categories. The first category uses quality assessment results or incremental-gain thresholds as stopping criteria. For example, RBQE~\cite{xing2020early} introduces an image quality assessment module for compressed image enhancement to determine whether the enhancement process should be terminated early. Adaptive Patch Exiting~\cite{wang2022adaptive}, designed for image super-resolution, determines the exit depth of each patch according to its incremental restoration gain at different network layers. The second category first estimates the restoration difficulty of input samples or local regions, and then assigns computational paths with different complexities. For example, ClassSR~\cite{kong2021classsr} divides image patches into easy, medium, and hard categories and sends them to SR branches with different capacities. CRESNet~\cite{chen2022blind} selects different restoration depths according to the compression severity to achieve adaptive computation for blind JPEG artifact removal. Although these methods have demonstrated the effectiveness of difficulty-aware computation allocation, they still have certain limitations. Threshold-based exiting methods usually require predefined stopping thresholds and introduce additional quality assessment or gain prediction modules during inference. Path-selection-based methods mainly focus on allocating different computational budgets to different samples or regions, but rarely further exploit fine-grained degradation representations to explicitly modulate enhancement features.

\noindent
\textbf{Degradation Prior Estimation.}
Existing methods mainly provide prior guidance for compressed image or video enhancement through explicit compression-parameter estimation, implicit degradation representation learning, or structured prior modeling. FBCNN~\cite{jiang2021towards} predicts an adjustable quality factor from compressed images and uses it to control the JPEG artifact removal process. BQEV~\cite{ding2023blind} predicts the quality of target frames and adaptively fuses the outputs of different artifact reduction modules for blind compressed video enhancement. InvCISR~\cite{guo2026compressed} addresses compressed image super-resolution through invertible degradation-restoration modeling, allowing degradation information to be preserved in the invertible structure and reused for subsequent restoration. HFUR~\cite{zhang2025hierarchical} exploits the frequency-domain origin of HEVC compression artifacts, derives DCT-domain priors through an implicit DCT transform, and designs ImpFreqUp and HIR modules for hierarchical frequency-based upsampling and iterative refinement, demonstrating the effectiveness of frequency-domain priors in HEVC compressed video enhancement. Although these methods introduce degradation priors from different perspectives, they mostly use them as single guidance signals, leaving further room to explore the multi-scale spatial structure information and global degradation semantics contained in the prior generation process.

\begin{figure*}[!t]
\centerline{\includegraphics[width=0.9\linewidth]{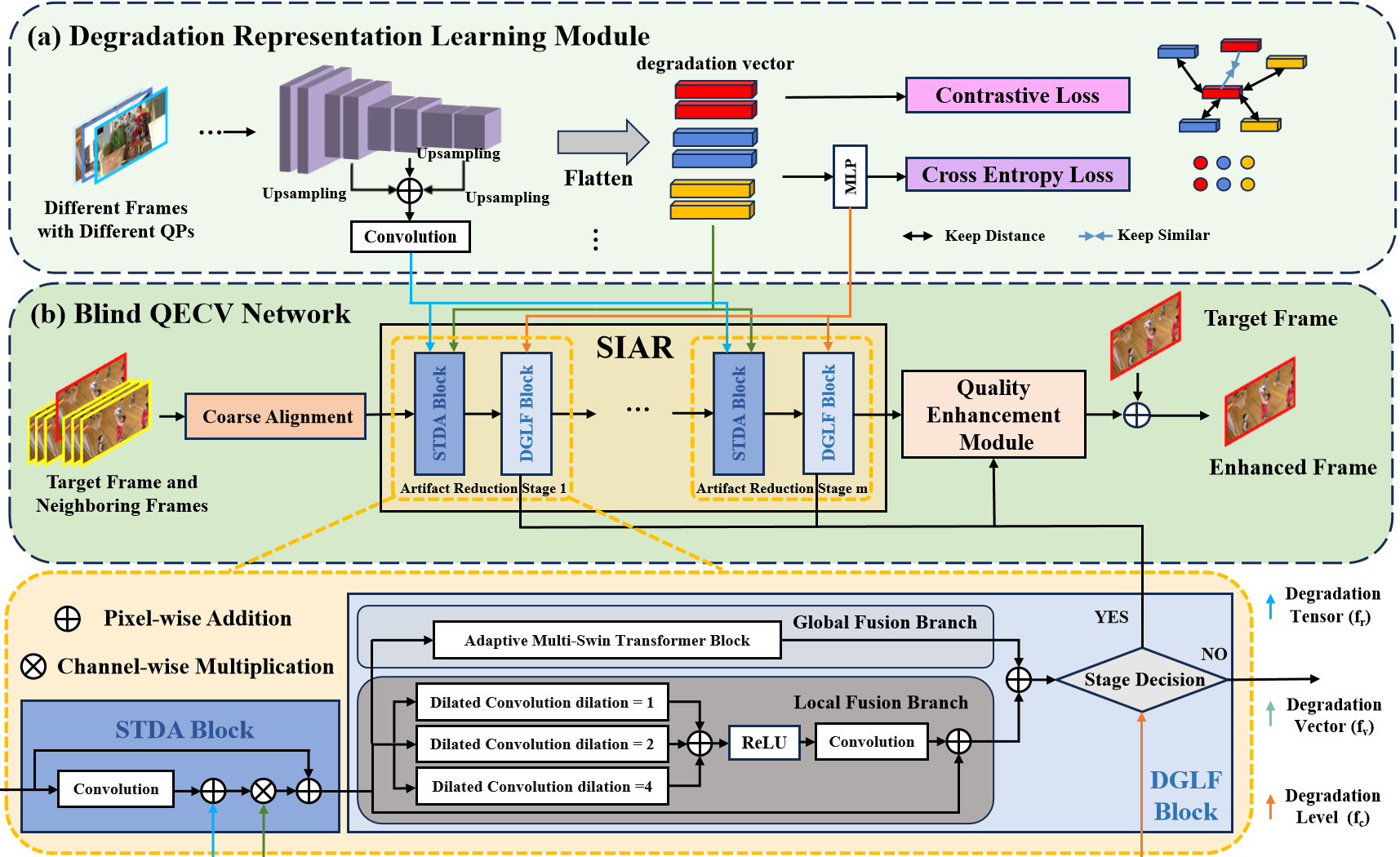}}

\caption{
The framework of proposed method, which comprises (a) Degradation Representation Learning (DRL) module and (b) blind QECV network. The DRL module extracts multi-scale degradation information of the target frame, including degradation tensor (blue arrow), degradation vector (green arrow), and degradation level (orange arrow), which is then fed into the blind QECV network. This network includes three key components: a Coarse Alignment module for frame alignment, a Sequential Inference-based Artifact Reduction (SIAR) module, and a Quality Enhancement module. The SIAR module incorporates up to five artifact reduction stages, each consisting of an STDA block and a DGLF block, enabling adaptive computational cost and efficient artifact removal based on degradation severity. Finally, the Quality Enhancement module further refines spatial features of the target frame to improve visual quality.}
\label{fig:overview}
\end{figure*}

\section{Method}

\subsection{Overview}

We propose a blind quality enhancement network for compressed video via fine-grained degradation-guided sequential inference. As shown in Fig.\ref{fig:overview}, the network consists of two parts: DRL module and blind QECV network that contains Coarse Alignment module, Sequential Inference-based Artifact Reduction (SIAR) module and Quality Enhancement (QE) module.

To learn the degradation representation and level of the compressed frames, the target frame is first fed into the DRL module, which utilizes an encoder structure. The learned degradation representations and level are then sent to the blind QECV network. To fully leverage the spatio-temporal information of the video, both the target frame $X_{t}$ at time $t$ and its neighboring frames are used as input. The compressed frame sequence with a radius of $r$ can be represented as $X = \{ X_{t-r}, \cdots, X_{t}, \cdots, X_{t+r} \}$. Initially, in the Coarse Alignment module, a deformable convolution \cite{deng2020spatio} is applied to align the input frames. The aligned spatio-temporal feature information is subsequently fed into the SIAR module. Within the SIAR module, the spatio-temporal features are aggregated with the degradation representation through artifact reduction stages for feature enhancement. Additionally, the degradation level predicted by the DRL module determines whether sequential inference should stop at the current artifact reduction stage. If not, the features are passed to the next artifact reduction stage. If yes, they would be sent to the QE module \cite{yu2024multi}, which selects useful channels at different scales to exploit spatial information to obtain the enhanced frame. The DRL module employs a pretraining strategy. When training the blind QECV network, the parameters of the DRL module would be frozen. The details of the DRL module, blind QECV network are shown in the following sections.

\subsection{Degradation Representation Learning Module}

To extract high-dimensional fine-grained degradation features, we design a DRL module that explicitly encodes degradation patterns typically entangled with image content. Serving as a preparatory module for the blind QECV network, this module helps decouple content and degradation information, providing structured degradation priors that enhance the overall performance of the enhancement process. 
Given the input frame $X_t$, it is passed through a hierarchical encoder comprising four sequential stages. Each encoder stage $E_i(\cdot)$ is constructed with stacked residual blocks followed by spatial downsampling. In this way, 
The encoder features $f_0, f_1, f_2, f_3$ are recursively computed as:
\begin{equation}
f_i = E_i(f_{i-1}),\quad  i \in \{0, 1, 2, 3\},
\label{eq:recursive_encoding}
\end{equation}
where $f_{-1}$ denotes the input frame $X_{t}$.

This recursive structure enables hierarchical abstraction of degradation cues at multiple resolutions.
To capture more stable degradation features, the last three encoder outputs $f_1, f_2, f_3$ are bilinearly upsampled, element-wise summed, and passed through a convolutional layer to generate the degradation tensor:
\begin{equation}
f_r = \text{Conv}\left(\text{Up}(f_1) + \text{Up}(f_2) + \text{Up}(f_3)\right),
\label{eq:degradation_tensor_fused}
\end{equation}
where $\text{Up}(\cdot)$ denotes bilinear interpolation. The resulting degradation tensor $f_r$ preserves spatial structural information and serves as pixel-wise guidance for subsequent enhancement stages.
The final encoder output $f_3$ is first flattened into a degradation vector $f_v$, which is then mapped to a discrete degradation class $f_c$ through a multi-layer perceptron:
{\small
\begin{equation}
f_v = \text{Flatten}(f_3)
\end{equation}
\begin{equation}
f_c = \text{MLP}(f_v) 
\end{equation}}
This process compactly encodes the overall degradation characteristics of one frame. The predicted degradation level $f_c$ serves a dual purpose: it provides a conditioning prior to the blind QECV network and also supports auxiliary classification loss during training.

For training, $N$ different single frames with varying QP values are randomly selected. The input frames are randomly cropped into a total of $2N$ patches, which are used as a mini-batch input to the encoder network. For the outputs of the encoder, the degradation vectors are utilized to calculate contrastive loss, and the degradation levels are employed to calculate classification loss. For classification loss, cross-entropy loss~\cite{pmlr-v202-mao23b} is employed, which can be formulated as:
\begin{equation}
L_{CrossEntropy} = -\frac{1}{2N}\sum_{i=1}^{2N}\sum_{c=1}^{C} 
\hat{f}^{\,i}_c \,\log\!\big(f^{\,i}_c\big),
\end{equation}
where $f^{\,i}_c$ denotes the predicted probability of class $c$ for the $i$-th sample, and $\hat{f}^{\,i}_c$ is the corresponding one-hot ground-truth label.

For the contrastive loss, the InfoNCE loss function~\cite{InfoNCE} is used, where patches from the same frame are positive samples and patches from the different frames are negative samples. The InfoNCE loss function can be defined as follows:
{\small
\begin{equation}
L_{InfoNCE}
= - \frac{1}{2N} \sum_{i=1}^{2N}
\log \frac{\exp\!\big(\mathrm{sim}(f_v^{i}, f_v^{i^+})/\tau\big)}
{\sum_{\substack{k=1,\,k\neq i}}^{2N} \exp\!\big(\mathrm{sim}(f_v^{i}, f_v^{k})/\tau\big)},
\end{equation}
where $f_v^i$ represents the degradation vector of the i-th sample, $f_v^{i^+}$ is its positive counterpart in the mini-batch, $\tau$ is a temperature parameter, and $\text{sim}(\cdot,\cdot)$ denotes cosine similarity.}

The overall loss function $L$ for the DRL is:
\begin{equation}
L_{total} = L_{CrossEntropy} + \lambda L_{InfoNCE},
\end{equation}
where $\lambda$ is a hyper parameter that balances the influence of two losses.

\subsection{Blind QECV Network}
As shown in Fig. \ref{fig:overview}, the proposed blind QECV network consists of three stages: Coarse Alignment (CA), Sequential Inference-based Artifact Reduction (SIAR), and Quality Enhancement (QE). It takes as input a sequence of $2r + 1$ frames $X$, fine-grained degradation representations (including the degradation tensor $f_r$ and the degradation vector $f_v$), and a degradation level $f_c$. The CA module aligns temporal motion using deformable convolution and outputs the initial spatio-temporal feature $f^0_{st}$. The SIAR module then performs adaptive artifact reduction guided by $f_r$, $f_v$, and $f_c$. It contains multiple artifact reduction stages, each composed of a STDA (Spatial-Temporal Information and Degradation
Representation Aggregation) block and a DGLF (Dual Global and Local Fusion) block. STDA aggregates $f_{st}$ with $f_r$ and $f_v$ via feature addition and channel attention. DGLF adopts a dual-branch design: one branch uses Multi-Swin Transformer~\cite{yu2024multi} for global modeling, and the other applies dilated convolution for local detail. To balance performance and efficiency, a sequential inference strategy is applied, where the number of artifact reduction stages is determined by $f_c$. The output is finally passed to the QE module, a lightweight cascaded reconstruction network that progressively refines compressed features through channel-attention-based selective fusion. The details are as follows:
\vspace{2pt}

\noindent
\textbf{STDA Block.}
As shown in Fig. \ref{fig:overview}, the input feature $f^{i-1}_{st}$ at i-th STDA block is first enhanced via a $3 \times 3$ convolution. The enhanced feature is then combined with the degradation tensor $f_r$ by element-wise addition. The combined features are multiplied channel-wisely with the degradation vector $f_v$, which is used as a feature weighting factor. The aggregated information is then added as a residual to $f^{i-1}_{st}$ to produce the output. The entire process can be represented as:
\begin{equation}
f^i_a = f^{i-1}_{st} + ((Conv(f^{i-1}_{st}) + f_r) \times f_v) ,
\end{equation}
where $f^i_a$ is the aggregated features at i-th STDA block.

\vspace{3pt}

\noindent
\textbf{DGLF Block.}
Inspired by the PSCF block~\cite{zhang2023video}, the DGLF module adopts a dual-branch architecture that fuses global and local features. As shown in Fig. \ref{fig:overview}, the aggregated features $f^i_a$ are fed into a multi-Swin Transformer-based global branch to better perceive spatio-temporal information by capturing long-range dependencies, and into a local branch with multi-scale dilated convolutions to recover high-frequency details and adapt to block-based compression. Both branches use residual connections, and their outputs are fused by element-wise addition. The dual-branch fusion process can be formulated as:

\begin{equation}
f_{g}^i = MSwin(f_{a}^{i}),
\end{equation}

\begin{equation}
f_{l}^i = Conv(\mathrm{ReLU}(\sum DConv_d(f_{a}^{i}))), d = \{1, 2, 4\},
\end{equation}

\begin{equation}
f_{st}^i = f_{g}^i + f_{l}^i ,
\end{equation}
where $f_{g}^i$, $f_{l}^i$, $f_{st}^i$ represent the outputs of the global fusion branch, the local fusion branch, and the artifact reduction stage, respectively, at stage $i$. Meanwhile, $MSwin()$ represents the multi-Swin Transformer, ReLU represents the ReLU activation function, and $DConv_d()$ represents dilation convolution with dilation rates $d$. For the stage decision process, if the index of current artifact reduction stage equals to the degradation level, the spatial-temporal feature $f_{st}^i$ would be sent to the QE module proposed in~\cite{yu2024multi}. Otherwise, it would be fed into next artifact reduction stage.

For training, we use Charbonnier Loss to optimize the parameters. The loss function is defined as:
\begin{equation}\begin{array}
{rcl}{L}_{charb} & = & \sqrt{\left(X_{t}^{e}-X_{t}^{raw}\right)^{2}+\epsilon},
\end{array}\end{equation}
where $X_{t}^{e}$ represents enhanced frame, $X_{t}^{raw}$ represents raw frame, and  $\epsilon$ is a constant set to $10^{-6}$.

\begin{table*}
\centering
\small
\caption{Overall comparison for $\Delta$ PSNR and $\Delta$ SSIM ($\times 10^{-2}$) on HEVC dataset at five QPs.}
\setlength{\tabcolsep}{2pt}
\begin{tabular}{cc|ccc|ccc|ccc|ccc|ccc}
\hline
\multicolumn{2}{c|}{\multirow{2}{*}{Methods}}              & \multicolumn{3}{c|}{QP22}                                        & \multicolumn{3}{c|}{QP27}                                        & \multicolumn{3}{c|}{QP32}                                        & \multicolumn{3}{c|}{QP37}                                        & \multicolumn{3}{c}{QP42}                                        \\ \cline{3-17} 
\multicolumn{2}{c|}{}                                      & PSNR          & SSIM                  & Time                     & PSNR          & SSIM                  & Time                     & PSNR          & SSIM                  & Time                     & PSNR          & SSIM                  & Time                     & PSNR          & SSIM                  & Time                    \\ \hline
\multicolumn{1}{c|}{\multirow{4}{*}{Non-Blind}} & STDF-R3L & 0.63          & 0.34                  & \textbf{0.4h}                         & 0.72          & 0.57                  & \textbf{0.4h}                         & 0.86          & 1.04                  & \textbf{0.4h}                         & 0.83          & 1.51                  & \textbf{0.4h}                         & 0.76          & 2.04                  & \textbf{0.4h}                        \\
\multicolumn{1}{c|}{}                           & RFDA     & 0.76          & 0.42                  & \underline{0.5h}                      & 0.82          & 0.68                  & \underline{0.5h}                      & 0.87          & 1.07                  & \underline{0.5h}                      & 0.91          & 1.62                  & \underline{0.5h}                      & 0.82          & 2.20                  & \underline{0.5h}                     \\
\multicolumn{1}{c|}{}                           & STDR     & \textbf{0.87} & \textbf{0.48}         & -                        & \textbf{0.97} & \underline{0.81}                  &  -                       & \underline{0.99}          & 1.24                  & -                        & 0.98          & 1.79                  & -                        & \underline{0.95}          & \underline{2.47}                  & -                       \\
\multicolumn{1}{l|}{}                           & M-Swin-T   & \underline{0.85}          & \textbf{0.48}         & 0.7h                      & 0.96          & \textbf{0.82}         & 0.7h                      & \textbf{1.01} & \textbf{1.30}         & 0.7h                      & \underline{1.01}          & \underline{1.83}                  & 0.7h                      & 0.91          & 2.46                  & 0.7h                     \\ \hline
\multicolumn{1}{c|}{\multirow{4}{*}{Blind}}     & FBCNN    & 0.29          & 0.19                  & \multicolumn{1}{c|}{1.1h}                         & 0.39          & 0.38                  & \multicolumn{1}{c|}{1.1h}                         & 0.42          & 0.60                  & \multicolumn{1}{c|}{1.1h}                         & 0.45          & 0.94                  & \multicolumn{1}{c|}{1.1h}                         & 0.47          & 1.55                  & \multicolumn{1}{c}{1.1h}                        \\
\multicolumn{1}{c|}{}                           & CRESNet  & 0.33          & 0.21                  & \multicolumn{1}{c|}{0.5h} & 0.40          & 0.39                  & \multicolumn{1}{c|}{0.5h} & 0.44          & 0.60                  & \multicolumn{1}{c|}{0.6h} & 0.49          & 1.00                  & \multicolumn{1}{c|}{0.8h} & 0.49          & 1.60                  & \multicolumn{1}{c}{0.9h} \\
\multicolumn{1}{c|}{}                           & BQEV     & 0.31          & \multicolumn{1}{c}{-} & \multicolumn{1}{c|}{-}   & 0.46          & \multicolumn{1}{c}{-} & \multicolumn{1}{c|}{-}   & 0.56          & \multicolumn{1}{c}{-} & \multicolumn{1}{c|}{-}   & 0.65          & \multicolumn{1}{c}{-} & \multicolumn{1}{c|}{-}   & 0.53          & \multicolumn{1}{c}{-} & \multicolumn{1}{c}{-}   \\
\multicolumn{1}{c|}{}                           & Ours     & 0.65          & \underline{0.40}                  & \multicolumn{1}{c|}{0.6h} & \underline{0.90}          & 0.78                  & \multicolumn{1}{c|}{0.8h} & \textbf{1.01} & \underline{1.27}                  & \multicolumn{1}{c|}{0.9h} & \textbf{1.03} & \textbf{1.88}         & \multicolumn{1}{c|}{1.0h} & \textbf{0.98} & \textbf{2.56}         & \multicolumn{1}{c}{1.2h} \\ \hline
\end{tabular}
\label{tab:HEVC}
\end{table*}

\section{Experiments}
\subsection{Setups}

\noindent
\textbf{Dataset.}
We use the MFQEv2 dataset~\cite{guan2019mfqe}, which provides training and testing videos at various resolutions. Following the standard setting, compressed videos are generated using HEVC~\cite{sullivan2012overview} and VVC~\cite{bross2021overview}. The models are trained on five seen QPs (22, 27, 32, 37, 42) and evaluated on both seen and unseen QPs. In addition, to assess cross-dataset generalization, we further test our model on the Vimeo-90K dataset~\cite{xue2019video}. Specifically, we select its clean test sequences and compress them with 5 QPs, and then use the resulting compressed sequences for evaluation.

\noindent
\textbf{Implementation Details.}
All experiments were conducted on a server equipped with an NVIDIA Tesla V100S-PCIE-32GB GPU and Intel Xeon Gold 6240 CPUs, using CUDA 11.7. During training, 128×128 patches are randomly sampled from consecutive frames with random flipping and rotation for data augmentation. We use the Adam optimizer~\cite{kingma2014adam} with an initial learning rate of 1 $\times$ $10^{-4}$ and a fixed 3×3 kernel size. The batch size is set to 32. For DRL pretraining, the encoder uses a downsampling scale of 2 and channel dimensions of [64, 64, 128, 256]. The SIAR module consists of 5 artifact reduction stages, corresponding to the 5 discrete degradation levels (QPs) defined in the standard MFQEv2 setting.

\subsection{Comparison with State-of-the-Art Methods}
In order to verify the superior performance of the proposed model, we select the SOTA non-blind QECV methods STDF-R3L~\cite{deng2020spatio}, RFDA~\cite{zhao2021recursive}, STDR~\cite{luo2022spatio}, M-Swin~\cite{yu2024multi}, and blind QECV methods, including FBCNN~\cite{jiang2021towards}, CRESNet~\cite{chen2022blind} and BVQE~\cite{ding2023blind} for comparison. The experimental results of BQEV were obtained from its original paper as its source code is not publicly available.

\noindent
\textbf{Quantitative Comparison on HEVC.}
As shown in Table~\ref{tab:HEVC}, our method achieves the best PSNR at QP32, QP37, and QP42, and ranks second at QP27, only behind the non-blind STDR. For SSIM, it leads at QP37 and QP42, and comes close behind M-Swin-T at QP22 and QP32. At low-to-mid QPs (22–32), it remains among the top non-blind performers, while excelling over all at high QPs.

In terms of efficiency, non-blind methods (e.g., STDF-R3L, RFDA) are faster due to the absence of degradation inference. Among blind methods, our approach offers superior PSNR, SSIM, and speed compared to FBCNN at QP22–37. Though its average inference time is 27\% longer than CRESNet, it achieves over 2× higher PSNR. With the sequential inference strategy, inference time scales with degradation level, e.g., QP22 takes only half as long as QP42 (0.6h vs. 1.2h). Overall, our method strikes a strong balance between quality and efficiency.

\noindent
\textbf{Quantitative Comparison on VVC.}
As shown in Table~\ref{tab:VVC}, our method achieves optimal performance across all QPs and all video sequences, demonstrating the effectiveness of our method over VVC compressed videos. With QPs increasing (i.e. quality degrading), our method achieves more significant quality improvements in both PSNR and SSIM, with 0.5 dB difference in PSNR and 0.01 in SSIM. In contrast, FBCNN and CRESNet maintain similar gains over all QPs. This demonstrates our method is able to recognize the QP level (i.e. degradation representation and level) and enhance the quality accordingly. Besides, for the challenging BQTerrace sequence, our method achieves PSNR improvement of 0.33 dB, beating the second-best result with a huge gain of 0.07 dB.
This result validates the performance of our method across diverse video content, especially for challenging ones. 

\begin{table}[H]
\centering
\small
\caption{$\Delta$ PSNR and $\Delta$ SSIM ($\times 10^{-2}$) comparison of blind methods on VVC dataset over 18 test sequences.}
\setlength{\tabcolsep}{1pt}
\begin{tabular}{c|c|c|cc|cc|cc}
\hline
\multirow{2}{*}{QP} & \multirow{2}{*}{Class} & \multirow{2}{*}{Sequence} 
& \multicolumn{2}{c|}{FBCNN} 
& \multicolumn{2}{c|}{CRESNet} 
& \multicolumn{2}{c}{Ours} \\ 
\cline{4-9}
& & & PSNR & SSIM & PSNR & SSIM & PSNR & SSIM \\
\hline
\multirow{19}{*}{37} & \multicolumn{1}{c|}{\multirow{2}{*}{A}}     & Traffic                   & 0.13        & 0.35        & \underline{0.14}   & \underline{0.38}   & \textbf{0.59} & \textbf{0.88} \\
                     & \multicolumn{1}{c|}{}                       & PeopleOnStreet            & \underline{0.10}  & \underline{0.27}  & 0.02         & 0.16         & \textbf{0.82} & \textbf{1.33} \\ \cline{2-9} 
                     & \multicolumn{1}{c|}{}                       & Kimino                    & \underline{ 0.12}  & \underline{0.34}  & 0.09         & 0.31         & \textbf{0.79} & \textbf{1.33} \\
                     & \multicolumn{1}{c|}{}                       & ParkScene                 & \underline{0.08}   & 0.38     & \underline{0.08}   & \underline{0.43}   & \textbf{0.55} & \textbf{1.47} \\
                     & \multicolumn{1}{c|}{B}                      & Cactus                    & \underline{0.11}  & \underline{0.31}  & 0.06         & 0.26         & \textbf{0.54} & \textbf{1.06} \\
                     & \multicolumn{1}{c|}{}                       & BQTerrace                 & \underline{0.07}  & \underline{0.21}  & 0.04         & 0.19         & \textbf{0.33} & \textbf{0.58} \\
                     & \multicolumn{1}{c|}{}                       & BasketballDrive           & \underline{0.15}  & \underline{0.37}  & 0.05         & 0.18         & \textbf{0.56} & \textbf{0.86} \\ \cline{2-9} 
                     & \multicolumn{1}{c|}{\multirow{4}{*}{C}}     & RaceHorses                & \underline{0.13}  & \underline{0.47}  & 0.08         & 0.33         & \textbf{0.39} & \textbf{1.13} \\
                     & \multicolumn{1}{c|}{}                       & BQMall                    & \underline{0.19}  & \underline{0.53}  & 0.10         & 0.37         & \textbf{0.84} & \textbf{1.49} \\
                     & \multicolumn{1}{c|}{}                       & PartyScene                & \underline{0.12}  & \underline{0.44}  & 0.11         & 0.34         & \textbf{0.44} & \textbf{1.33} \\
                     & \multicolumn{1}{c|}{}                       & BasketballDrill           & \underline{0.17}  & \underline{0.44}  & 0.16          
                     & \underline{0.44}   & \textbf{0.44} & \textbf{0.79} \\ \cline{2-9} 
                     & \multicolumn{1}{c|}{\multirow{4}{*}{D}}     & RaceHorses                & \underline{0.14}  & \underline{0.49}  & 0.12         & 0.37         & \textbf{0.66} & \textbf{1.98} \\
                     & \multicolumn{1}{c|}{}                       & BQSquare                  & 0.18        & 0.42        & \underline{0.23}   
                     & \underline{0.49}   & \textbf{0.61} & \textbf{0.89} \\
                     & \multicolumn{1}{c|}{}                       & BlowingBubbles            & \underline{0.14}  & \underline{0.62}  & 0.13         & 0.55         & \textbf{0.54} & \textbf{1.84} \\
                     & \multicolumn{1}{c|}{}                       & BasketballPass            & \underline{0.21}  & \underline{0.67}  & 0.18         & 0.52         & \textbf{0.91} & \textbf{2.07} \\ \cline{2-9} 
                     & \multicolumn{1}{c|}{}                       & FourPeople                & \underline{0.21}  & \underline{0.37}  & 0.13         & 0.29         & \textbf{0.66} & \textbf{0.71} \\
                     & \multicolumn{1}{c|}{E}                      & Johnny                    & \underline{0.19}  & \underline{0.25}  & 0.15         & 0.19         & \textbf{0.50} & \textbf{0.40} \\
                     & \multicolumn{1}{c|}{}                       & KristenAndSara            & \underline{0.19}  & \underline{0.29}  & 0.09         & 0.20         & \textbf{0.65} & \textbf{0.56} \\ \cline{2-9} 
                     & \multicolumn{2}{c|}{Average}                                            & \underline{0.15}  & \underline{0.40}  & 0.11         & 0.33         & \textbf{0.60} & \textbf{1.15} \\ \hline
22                   & \multicolumn{2}{c|}{Average}                                            & \underline{0.09}  & \underline{0.09}  & 0.07         & 0.06         & \textbf{0.36} & \textbf{0.23} \\ \hline
27                   & \multicolumn{2}{c|}{Average}                                            & \underline{0.12}  & \underline{0.17}  & 0.10         & 0.14         & \textbf{0.23} & \textbf{0.27} \\ \hline
32                   & \multicolumn{2}{c|}{Average}                                            & \underline{0.14}  & \underline{0.28}  & 0.10         & 0.23         & \textbf{0.39} & \textbf{0.62} \\ \hline
42                   & \multicolumn{2}{c|}{Average}                                            & \underline{0.15}  & \underline{0.55}  & 0.12         & 0.47         & \textbf{0.51} & \textbf{1.18} \\ \hline
\end{tabular}
\label{tab:VVC}
\end{table}

\noindent
\textbf{Robustness to Unseen QPs Evaluation.}
To assess robustness to unseen QPs, we evaluate our method on QPs unseen during training. As shown in Table~\ref{tab:unselct}, our approach achieves the best PSNR and SSIM across all five unseen QPs compressed with HEVC and VVC. Except for QP30 in VVC, PSNR improvements exceed 100\% and SSIM gains are no less than 90\%. For instance, FBCNN even shows degradation at HEVC QP20 (-0.02dB, -0.05), while our method maintains consistent improvements.

Furthermore, performance differences between seen and unseen QPs are minimal. In HEVC, PSNR and SSIM at unseen QP35 are only 0.03dB and 0.27 lower than seen QP37 (3\% and 14\% drops). In VVC, the gaps are 0.06dB and 0.19 (10\% and 17\%). By contrast, FBCNN exhibits 0.31dB and 0.24 degradation at unseen QP20 vs. seen QP32 in VVC (107\% and 126\% declines). These results highlight the robustness of our method.

\begin{table*}[htbp]
\centering

\caption{$\Delta$ PSNR and $\Delta$ SSIM ($\times 10^{-2}$) comparison of blind methods on HEVC and VVC Datasets for unseen QPs.}
\begin{tabular}{cc|cc|cc|cc|cc|cc}
\hline
\multicolumn{2}{c|}{\multirow{2}{*}{Methods}} & \multicolumn{2}{c|}{QP20}                            & \multicolumn{2}{c|}{QP25}                            & \multicolumn{2}{c|}{QP30}                            & \multicolumn{2}{c|}{QP35}                            & \multicolumn{2}{c}{QP40}                            \\ \cline{3-12} 
\multicolumn{2}{c|}{}                         & \multicolumn{1}{l}{PSNR} & \multicolumn{1}{l|}{SSIM} & \multicolumn{1}{l}{PSNR} & \multicolumn{1}{l|}{SSIM} & \multicolumn{1}{l}{PSNR} & \multicolumn{1}{l|}{SSIM} & \multicolumn{1}{l}{PSNR} & \multicolumn{1}{l|}{SSIM} & \multicolumn{1}{l}{PSNR} & \multicolumn{1}{l}{SSIM} \\ \hline
\multicolumn{1}{c|}{}          & FBCNN        & -0.02                    & -0.05                     & \underline{0.38}               & \underline{0.31}                & 0.40                     & 0.49                      & 0.42                     & 0.74                      & 0.42                     & \underline{1.17}                     \\
\multicolumn{1}{c|}{HEVC}      & CRESNet      & \underline{0.25}                     & \underline{0.14}                      & \underline{0.38}               & \underline{0.31}                & \underline{0.43}               & \underline{0.51}                & \underline{0.47}                     & \underline{0.80}                      & \underline{0.47}                     & 0.24                     \\
\multicolumn{1}{c|}{}          & Ours         & \textbf{0.52}            & \textbf{0.28}             & \textbf{0.80}            & \textbf{0.60}             & \textbf{0.92}            & \textbf{1.00}             & \textbf{1.00}            & \textbf{1.61}             & \textbf{0.96}            & \textbf{2.22}            \\ \hline
\multicolumn{1}{c|}{}          & FBCNN        & \underline{0.07}                     & \underline{0.07}                      & \underline{0.11}                     & \underline{0.13}                      & \underline{0.14}                     & \underline{0.23}                      & \underline{0.14}                     & \underline{0.35}                      & \underline{0.15}                     & \underline{0.48}                     \\
\multicolumn{1}{c|}{VVC}       & CRESNet      & 0.05                     & 0.05                      & 0.09                     & 0.10                      & 0.10                     & 0.19                      & 0.11                     & 0.30                      & 0.12                     & 0.41                     \\
\multicolumn{1}{c|}{}          & Ours         & \textbf{0.34}               & \textbf{0.18}                & \textbf{0.30}                     & \textbf{0.26}                      & \textbf{0.26}                     & \textbf{0.40}                      & \textbf{0.54}               & \textbf{0.96}                & \textbf{0.57}               & \textbf{1.24}               \\ \hline
\end{tabular}
\label{tab:unselct}
\end{table*}

\begin{table*}[t]
\centering
\caption{Quantitative comparison on Vimeo-90K under different compression levels: $\Delta$PSNR and $\Delta$SSIM ($\times 10^{-2}$).}
\label{tab:vimeo_generalization}
\setlength{\tabcolsep}{6pt}
\renewcommand{\arraystretch}{1.15}
\small
\begin{tabular}{c cc|cc|cc|cc}
\toprule
\multirow{2}{*}{QP} 
& \multirow{2}{*}{Avg. Compressed PSNR} 
& \multirow{2}{*}{Avg. Compressed SSIM}
& \multicolumn{2}{c|}{FBCNN}
& \multicolumn{2}{c|}{CRESNet}
& \multicolumn{2}{c}{Ours} \\
\cmidrule(lr){4-5} \cmidrule(lr){6-7} \cmidrule(l){8-9}
& & & PSNR & SSIM & PSNR & SSIM & PSNR & SSIM \\
\midrule
22 & 42.44 & 0.98
   & \underline{0.58} & \underline{0.24}
   & 0.56 & 0.23
   & \textbf{1.13} & \textbf{0.40} \\

27 & 38.99 & 0.96
   & \underline{0.55} & \underline{0.39}
   & 0.53 & 0.37
   & \textbf{1.53} & \textbf{0.86} \\

32 & 35.74 & 0.94
   & \underline{0.51} & \underline{0.65}
   & 0.50 & 0.60
   & \textbf{1.78} & \textbf{1.59} \\

37 & 32.89 & 0.90
   & \underline{0.46} & \underline{0.94}
   & 0.46 & 0.90
   & \textbf{1.75} & \textbf{2.28} \\

42 & 30.21 & 0.84
   & 0.39 & 1.33
   & \underline{0.40} & \underline{1.34}
   & \textbf{1.64} & \textbf{2.94} \\
\bottomrule
\end{tabular}
\end{table*}

\noindent
\textbf{Cross-Dataset Generalization on Vimeo-90K.}
To further evaluate the cross-dataset generalization ability of different blind quality enhancement methods, we conduct experiments on Vimeo-90K under five compression levels. As shown in Table~\ref{tab:vimeo_generalization}, our method consistently achieves the best performance across all QPs in terms of both $\Delta$PSNR and $\Delta$SSIM. Compared with FBCNN and CRESNet, the proposed method yields substantially larger quality gains, demonstrating stronger robustness when transferred to a different dataset. In particular, the performance margin becomes more pronounced as the compression level increases. At QP42, our method improves PSNR by 1.64 dB and SSIM by 2.94$\times 10^{-2}$, significantly outperforming the competing methods. These results indicate that our method not only performs well on the training-domain data, but also generalizes effectively to unseen data distributions and severe compression artifacts.

\begin{figure*}
\centerline{\includegraphics[width=0.85\linewidth]{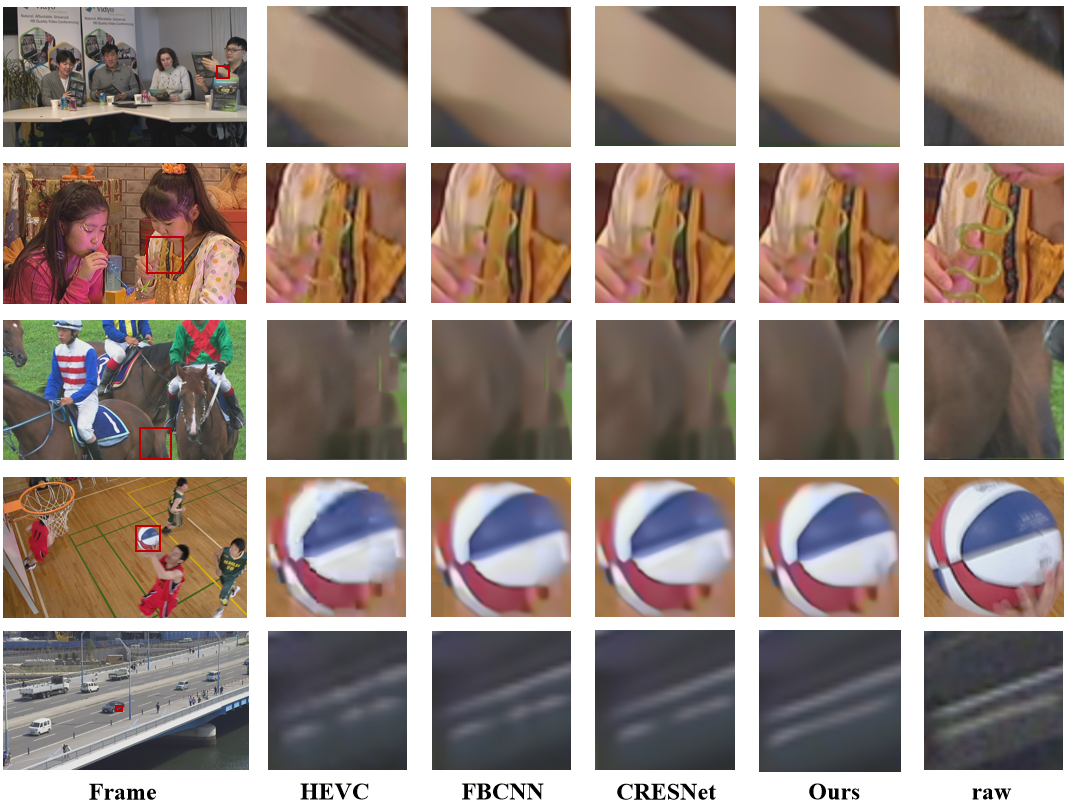}}
\caption{Detailed visualization on five sequences: BlowingBubbles (416x240), FourPeople (1280x720), RaceHorse(416x240), BasketballDrill (832x480), BQTerrace(1920x1080).}
\label{fig:visual}
\end{figure*}

\noindent
\textbf{Qualitative Comparison.}
The visualization results are shown in Fig. \ref{fig:visual}. In the first row, the two comparison methods produce overly smoothed results, causing the gloss on the straw to appear uniform. However, our method effectively restores the highlights on the straw. In the second row, the two comparison methods produce the same results in the hand region, causing the edges of the fingers blurry. In contrast, our method better preserves the natural skin texture and the subtle shading variations around the fingers. In the third row, while block artifacts from HEVC cause the horse’s tail and body to blend together and hinder accurate reconstruction by the comparison methods, our method removes these artifacts and vividly restores the tail’s streamlined texture. In the forth row, our method significantly reduces the blurriness around the basketball's edges, comparing to competing methods. In the last row, our method better restores the dynamic textures of the moving vehicle. In summary, our method outperforming other methods in challenges including: over-smoothing, loss of details, blocking artifacts, blurring and dynamic texture restoration.

\begin{table}
\centering
\caption{Model size, inference speed and performance with HEVC. Frame per second (FPS) and $\Delta$ PSNR (dB) are tested on all videos at five seen QPs.}

\begin{tabular}{c|ccccc}
\hline
Method   & Type      & Params(M)     & FPS           & PSNR          \\ \hline
STDF-R3L & Non-Blind & 1.27 $\times$ $n$                      & \textbf{5.54} & \underline{0.76}          \\
FBCNN    & Blind     & 71.91                                     & 2.01          & 0.41          \\
CRESNet  & Blind     & \textbf{4.6}                        & \underline{3.35}    & 0.43          \\
Ours     & Blind     & \underline{4.8}                      & 2.46          & \textbf{0.91} \\ \hline
\end{tabular}

\label{tab:complexity}
\end{table}

\begin{table}
\centering
\caption{TFLOPs at Different Resolutions (QP=37): 2560$\times$1600 (Class A), 1920$\times$1080 (Class B), 832$\times$480 (Class C), 416$\times$240 (Class D), 1280$\times$720 (Class E)}
\begin{tabularx}{\linewidth}{@{}c *{5}{Y}@{}}
\toprule
\multirow{2}{*}{Method} & \multicolumn{5}{c}{TFLOPs @ Different Resolutions (QP=37)} \\
\cmidrule(lr){2-6}
 & A & B & C & D & E \\
\midrule
FBCNN   & 22.76 & 11.52 & 2.22 & 0.55 & 5.12 \\
CRESNet & \underline{13.21} & \underline{6.74} & \underline{1.29} & \underline{0.34} & \underline{3.04} \\
Ours    & \textbf{6.43} & \textbf{3.21} & \textbf{0.62} & \textbf{0.15} & \textbf{1.45} \\
\bottomrule
\end{tabularx}

\label{tab:tflops}
\end{table}

\noindent
\textbf{Network complexity Comparison.}
Table~\ref{tab:complexity} quantifies model complexity in terms of parameter count and inference speed. For non-blind methods, a dedicated model must be trained for each QP; thus multiple models are required and the total parameter count increases proportionally with the number of QPs. By contrast, blind methods use a single model for all QPs, yielding substantially fewer parameters than non-blind counterparts. Among blind methods, our model ranks second to CRESNet in parameter count while achieving a 112 \% improvement in PSNR (from 0.76 to 0.91) over CRESNet. In terms of inference speed, non-blind methods are faster because they do not include the degradation-level estimation step; among blind methods, our average inference speed is second only to CRESNet while still delivering higher PSNR. Overall, our method strikes a balanced trade-off among parameter size, inference speed, and PSNR. In addition, Table~\ref{tab:tflops} compares TFLOPs across different resolutions (Classes A–E) at QP=37. Our method achieves the lowest computational complexity.  Across various resolutions, it reduces TFLOPs by an average of approximately 73.7 \% compared with FBCNN, while delivering an average TFLOPs reduction of around 50 \% relative to CRESNet, significantly outperforming existing blind methods.

\begin{figure}[htbp]
\centerline{\includegraphics[width=1.05\linewidth]{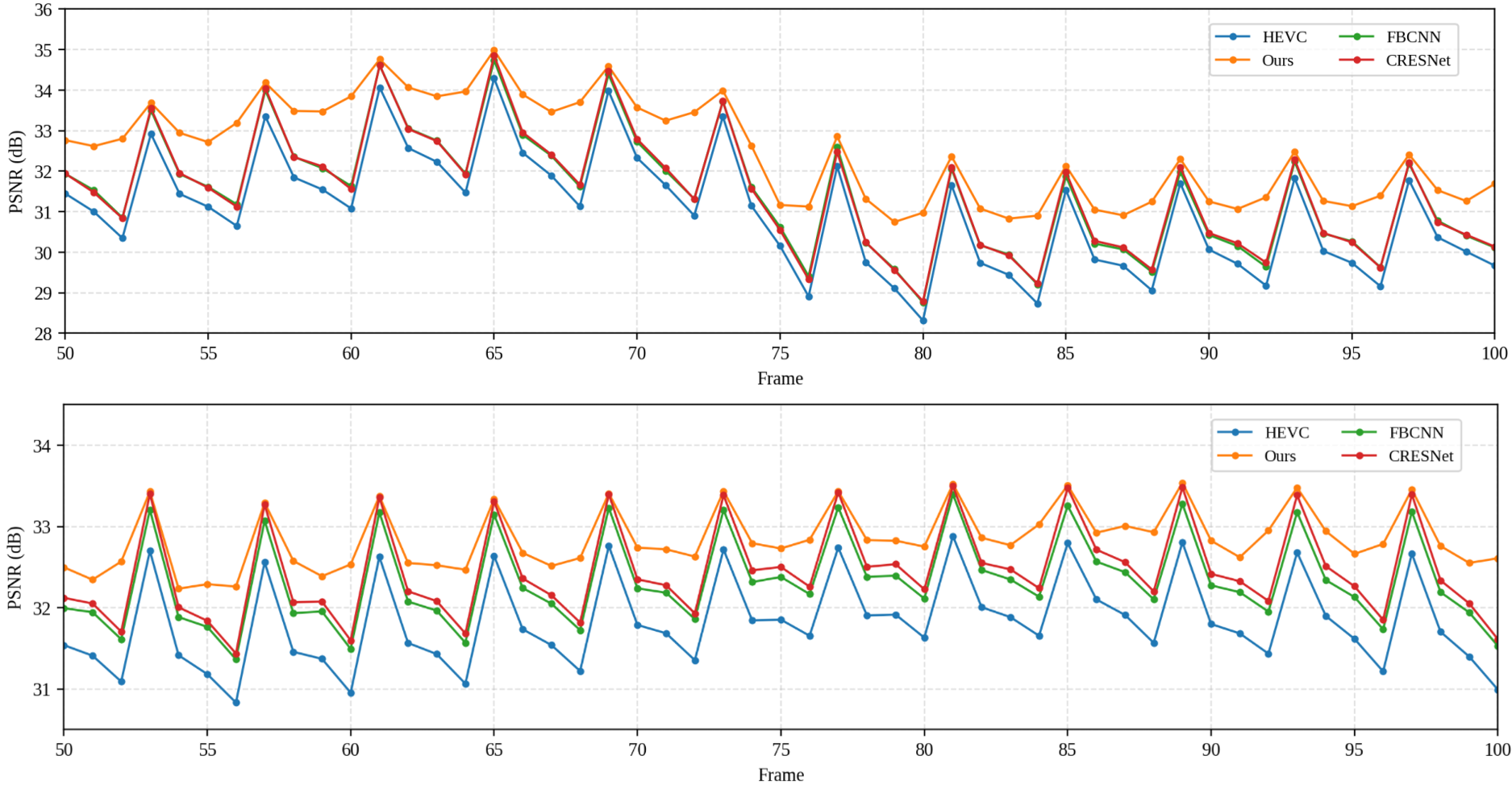}}
\caption{Illustration of quality fluctuations for two test sequences compressed with QP 37. (Top: Class D, BasketballPass. Bottom: Class C, BasketballDrill.)}
\label{fig:ypsnr}
\end{figure}

\noindent
\textbf{Quality Fluctuation.}
Frame-to-frame quality variations introduced during video compression or transmission disrupt visual continuity and severely degrade the user experience. As shown in Fig. \ref{fig:ypsnr}, we evaluate quality fluctuation by plotting per-frame PSNR curves for representative sequences. The results show that HEVC-compressed sequences and comparison methods such as FBCNN exhibit pronounced inter-frame fluctuations. In contrast, our method not only improves the overall PSNR but also significantly suppresses these fluctuations, producing smoother and more coherent video and thereby enhancing user experience.

\begin{figure}[htbp]
\centerline{\includegraphics[width=0.9\linewidth]{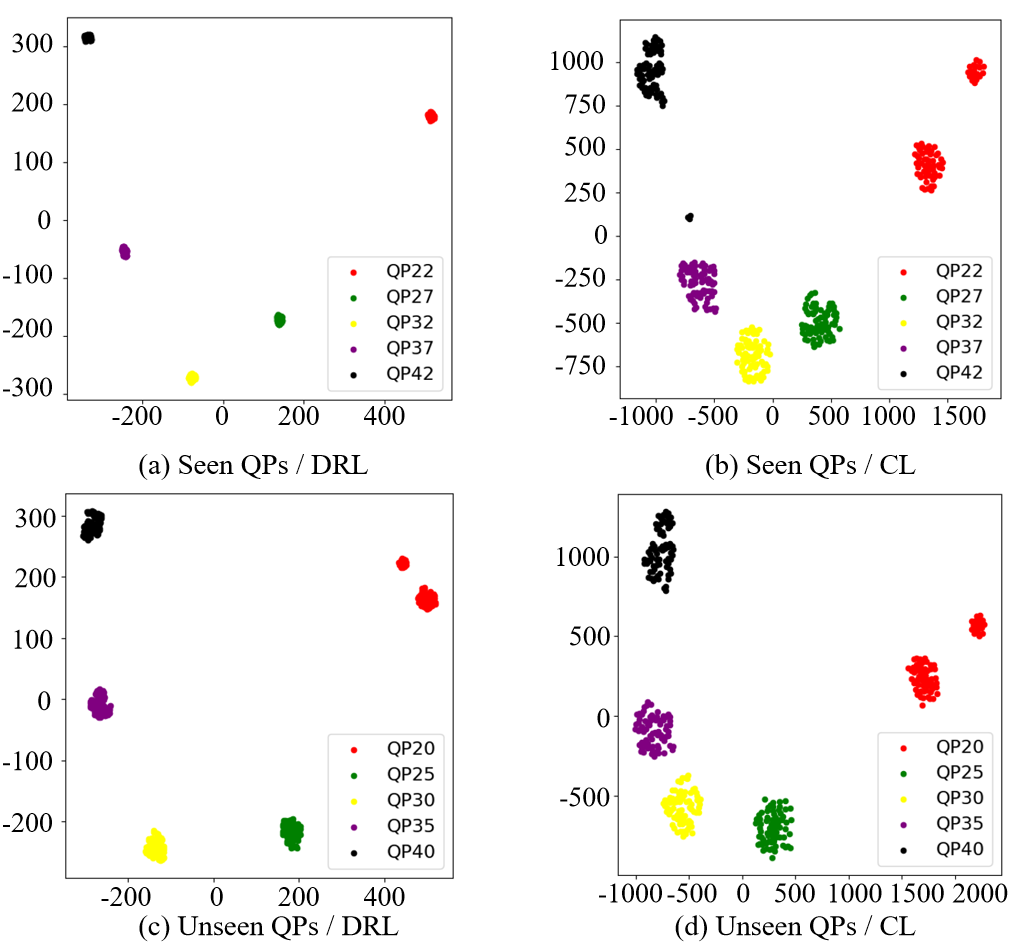}}
\caption{Visualization of Degradation Representation Learning (DRL) and Classification Learning (CL) on HEVC. (a) Clustering of DRL with seen QPs. (b) Clustering of DRL with unseen QPs. (c) Clustering of CL with seen QPs. (d) Clustering of CL with unseen QPs.}
\label{fig:rep_visual}
\end{figure}

\begin{figure}[!htbp]
    \centering
    \includegraphics[width=0.95\linewidth]{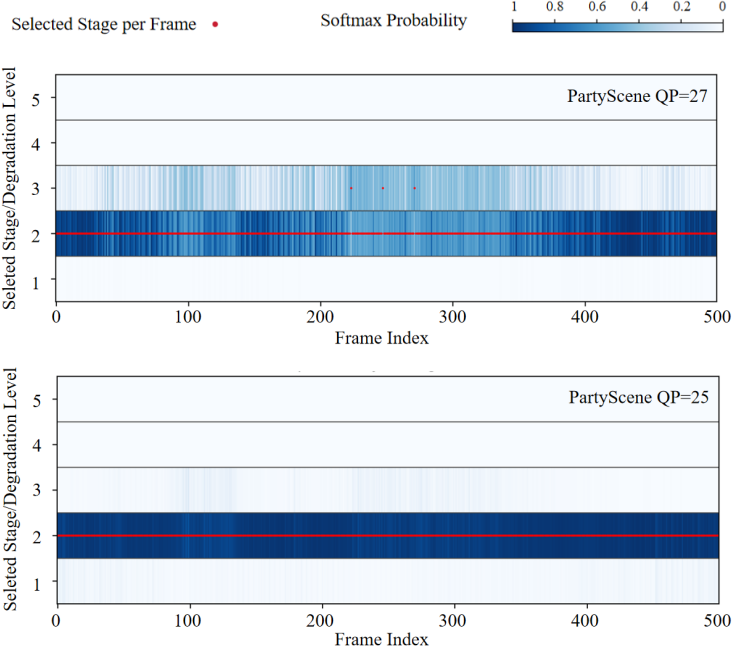}
    \vspace{0.5mm}
    
    \includegraphics[width=0.95\linewidth]{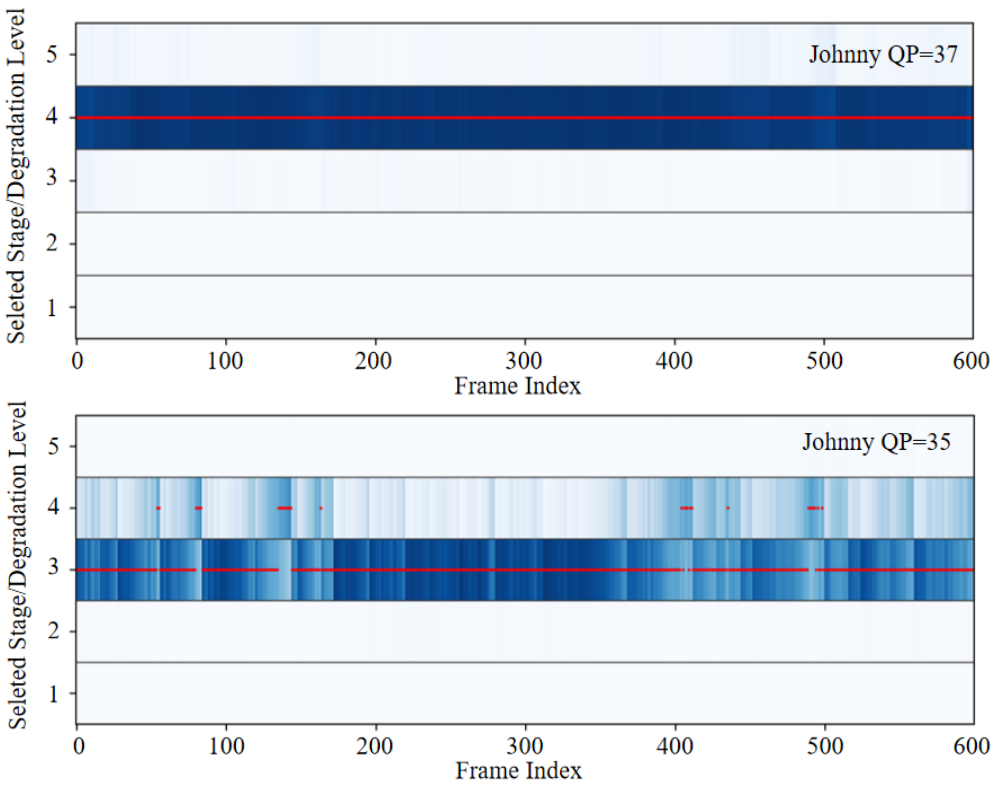}
\caption{Probability distribution of DRL-predicted degradation levels under seen (QP27\&37) and unseen (QP25\&35) QP settings. The horizontal axis denotes the frame index, and the vertical axis denotes the degradation level/stage index. The color changes from light to dark as the predicted softmax probability increases from 0 to 1, and the red dots indicate the selected artifact reduction stage for each frame.}
\label{fig:visual_stage}
\end{figure}

\begin{table}[!t]
\centering
\caption{degradation-content feature distance analysis.}
\label{tab:feature_distance}
\setlength{\tabcolsep}{12pt}
\begin{tabular}{c|ccc}
\hline
Method & $D_{\text{deg}}$ & $D_{\text{content}}$ & $\Delta$ \\
\hline
CE-only & 0.2321 & 0.0785 & 0.1536 \\
DRL     & 0.5112 & 0.2889 & \textbf{0.2223} \\
\hline
\end{tabular}
\end{table}

\subsection{DRL Prediction Distribution Analysis}
 To further analyze the routing behavior of the DRL module, we visualize the frame-level softmax probability distribution of the predicted degradation levels, as shown in Fig. \ref{fig:visual_stage}. For seen QPs, the predicted probabilities are highly concentrated around the corresponding degradation levels. For example, PartyScene QP27 is mainly assigned to level 2, and Johnny QP37 is mainly assigned to level 4, which is consistent with the training-stage mapping. For unseen QPs, the predictions are usually constrained within the local degradation range defined by neighboring training QPs. For example, QP35 lies between the training QPs of QP32 and QP37. As shown in Johnny QP35, most frames are predicted as level 3, while a small number of frames are predicted as level 4, corresponding to the two adjacent degradation levels of QP32 and QP37, respectively. This demonstrates the stability of DRL prediction.

\subsection{Visualization of DRL}
Fig. \ref{fig:rep_visual} shows the t-SNE visualizations of features extracted from the DRL and Classification Learning (CL) modules on both seen and unseen QPs. Compared to CL, DRL places more emphasis on degradation patterns, resulting in a more distinct clustering effect. In addition, we further conduct a degradation-content feature distance analysis to quantitatively examine the degradation-oriented property of the learned representation. As shown in Table~\ref{tab:feature_distance}, two types of feature distances are computed in the original degradation vector space. The first distance, denoted as $D_{\mathrm{deg}}$, represents the feature distance between the same frame from the same video sequence compressed with different QPs, reflecting the sensitivity of the representation to degradation variation. The second distance, denoted as $D_{\mathrm{content}}$, represents the feature distance between different video sequences compressed with the same QP, reflecting the sensitivity of the representation to content variation under the same degradation level. Furthermore, the separation margin $\Delta$ between these two distances is used to evaluate whether degradation variation is more distinguishable than content variation in the feature space. The results show that DRL increases $D_{\mathrm{deg}}$ from 0.2321 to 0.5112, indicating stronger sensitivity to compression degradation changes. Although $D_{\mathrm{content}}$ also increases, the separation margin $\Delta$ increases from 0.1536 to 0.2223. This indicates that, compared with classification-only learning, DRL more effectively enhances the feature separability caused by degradation variation.

\begin{table*}[t]
\centering
\renewcommand{\arraystretch}{1.15}
\setlength{\tabcolsep}{4.5pt}

\caption{Ablation study on DRL: $\Delta$PSNR and $\Delta$SSIM ($\times 10^{-2}$) at five QPs.}
\label{tab:ablation_drl}

\begin{tabular}{c|cc|cc|cc|cc|cc|cc}
\hline
\multicolumn{3}{c|}{DRL} 
& \multicolumn{2}{c|}{QP22} 
& \multicolumn{2}{c|}{QP27} 
& \multicolumn{2}{c|}{QP32} 
& \multicolumn{2}{c|}{QP37} 
& \multicolumn{2}{c}{QP42} \\ 
\hline

\multirow{2}{*}{Training Strategy} 
& \multicolumn{2}{c|}{Loss Function} 
& \multirow{2}{*}{PSNR} 
& \multirow{2}{*}{SSIM} 
& \multirow{2}{*}{PSNR} 
& \multirow{2}{*}{SSIM} 
& \multirow{2}{*}{PSNR} 
& \multirow{2}{*}{SSIM} 
& \multirow{2}{*}{PSNR} 
& \multirow{2}{*}{SSIM} 
& \multirow{2}{*}{PSNR} 
& \multirow{2}{*}{SSIM} \\ 
\cline{2-3}

& Classification 
& Contrastive 
&  
&  
&  
&  
&  
&  
&  
&  
&  
&  \\ 
\hline

\multirow{2}{*}{Frozen pretrained module} 
& \ding{51} 
& \ding{51} 
& \underline{0.65} 
& \textbf{0.40} 
& \textbf{0.90} 
& \textbf{0.78} 
& \textbf{1.01} 
& \textbf{1.27} 
& \textbf{1.03} 
& \textbf{1.88} 
& \textbf{0.98} 
& \textbf{2.56} \\

& \ding{51} 
& \ding{55} 
& \underline{0.65} 
& \underline{0.39} 
& \underline{0.87} 
& \underline{0.75} 
& \underline{0.96} 
& \underline{1.25} 
& 1.00 
& \underline{1.85} 
& \underline{0.96} 
& \textbf{2.56} \\ 
\cline{1-13}

Training from scratch 
& \ding{51} 
& \ding{51} 
& \textbf{0.67} 
& \underline{0.39} 
& 0.81 
& 0.71 
& 0.95 
& 1.20 
& \underline{1.01} 
& 1.83 
& 0.95 
& \underline{2.49} \\ 
\hline

\end{tabular}
\end{table*}

\begin{table*}[!t]
\centering
\caption{Ablation study on STDA: $\Delta$PSNR and $\Delta$SSIM ($\times 10^{-2}$) at five QPs.}
\begin{tabular}{c|ll|ll|ll|ll|ll}
\hline
\multirow{2}{*}{STDA} & \multicolumn{2}{c|}{QP22}     & \multicolumn{2}{c|}{QP27}     & \multicolumn{2}{c|}{QP32}     & \multicolumn{2}{c|}{QP37}     & \multicolumn{2}{c}{QP42}      \\ \cline{2-11} 
                      & PSNR          & SSIM          & PSNR          & SSIM          & PSNR          & SSIM          & PSNR          & SSIM          & PSNR          & SSIM          \\ \hline
Tensor+Vector         & \textbf{0.65} & \textbf{0.40} & \textbf{0.90} & \textbf{0.78} & \textbf{1.01} & \textbf{1.27} & \textbf{1.03} & \textbf{1.88} & \textbf{0.98} & \textbf{2.56} \\
Vector                & 0.61          & 0.38          & 0.85          & 0.73          & 0.96          & 1.22          & 0.99          & 1.81          & 0.96          & 2.53          \\ \hline
\end{tabular}
\label{tab:stda_ablation}
\end{table*}


\begin{table*}[]
\centering
\caption{Ablation study on Sequential Inference Strategy: $\Delta$ PSNR and $\Delta$ SSIM ($\times 10^{-2}$) at five QPs.}
\resizebox{0.95\linewidth}{!}{
\begin{tabular}{c|ccc|ccc|ccc|ccc|ccc}
\hline
\multirow{2}{*}{SIS} & \multicolumn{3}{c|}{QP22}                    & \multicolumn{3}{c|}{QP27}                    & \multicolumn{3}{c|}{QP32}                    & \multicolumn{3}{c|}{QP37}                    & \multicolumn{3}{c}{QP42}                     \\ \cline{2-16} 
                     & PSNR          & SSIM          & Time         & PSNR          & SSIM          & Time         & PSNR          & SSIM          & Time         & PSNR          & SSIM          & Time         & PSNR          & SSIM          & Time         \\ \hline
Over                 & \textbf{0.73} & \textbf{0.43} & 1.2h          & \textbf{0.90} & \textbf{0.78} & 1.2h          & 0.97          & 1.25          & 1.2h          & 1.00          & 1.84          & 1.2h          & \textbf{0.98} & \textbf{2.56} & \textbf{1.2h} \\
\ding{51}                    & 0.65          & 0.40          & \textbf{0.6h} & \textbf{0.90} & \textbf{0.78} & \textbf{0.8h} & \textbf{1.01} & \textbf{1.27} & \textbf{0.9h} & \textbf{1.03} & \textbf{1.88} & \textbf{1.0h} & \textbf{0.98} & \textbf{2.56} & \textbf{1.2h} \\
Under                & 0.65          & 0.40          & 0.6h          & 0.90          & 0.78          & 0.6h          & 0.86          & 1.14          & 0.6h          & 0.58          & 1.24          & 0.6h          & 0.31          & 1.06          & 0.6h          \\ \hline
\end{tabular}}
\label{tab:ablation}
\end{table*}


\begin{table*}[t]
\centering
    \caption{Comparison of adaptive computation mechanisms on the RaceHorses (416x240) sequence: $\Delta$ PSNR and $\Delta$ SSIM ($\times 10^{-2}$) at five QPs.}
\label{tab:exit_strategy_comparison}
\scriptsize
\setlength{\tabcolsep}{2.5pt}
\renewcommand{\arraystretch}{1.15}
\resizebox{\textwidth}{!}{
\begin{tabular}{c|ccc|ccc|ccc|ccc|ccc}
\hline
\multirow{2}{*}{Adaptive Computation Mechanism} 
& \multicolumn{3}{c|}{QP22} 
& \multicolumn{3}{c|}{QP27} 
& \multicolumn{3}{c|}{QP32} 
& \multicolumn{3}{c|}{QP37} 
& \multicolumn{3}{c}{QP42} \\ 
\cline{2-16}
& $\Delta$PSNR & $\Delta$SSIM & Time
& $\Delta$PSNR & $\Delta$SSIM & Time
& $\Delta$PSNR & $\Delta$SSIM & Time
& $\Delta$PSNR & $\Delta$SSIM & Time
& $\Delta$PSNR & $\Delta$SSIM & Time \\ 
\hline
IQA-based ($T_Q=0.90$)
& 0.64 & 0.41 & 69.28min
& 0.77 & 0.87 & 71.87min
& 0.85 & 1.54 & 74.17min
& 0.92 & 2.16 & 70.86min
& 0.93 & 2.70 & 71.24min \\

IQA-based ($T_Q=0.85$)
& 0.65 & 0.41 & 51.50min
& 0.77 & 0.86 & 59.09min
& 0.84 & 1.53 & 66.40min
& 0.93 & 2.16 & 71.17min
& 0.91 & 2.63 & 51.15min \\

Learnable-gate
& 0.80 & \textbf{0.50} & 9.16s
& \textbf{0.94} & \textbf{1.07} & 11.04s
& \textbf{0.99} & \textbf{1.93} & 14.12s
& \textbf{1.04} & \textbf{2.67} & 17.55s
& \textbf{1.04} & 3.26 & 20.39s \\

Sequential Inference Strategy (Ours)
& \textbf{0.81} & \textbf{0.50} & \textbf{5.95s}
& \textbf{0.94} & \textbf{1.07} & \textbf{6.86s}
& \textbf{0.99} & \textbf{1.93} & \textbf{8.37s}
& \textbf{1.04} & \textbf{2.67} & \textbf{9.80s}
& \textbf{1.04} & \textbf{3.27} & \textbf{11.34s} \\
\hline
\end{tabular}
}
\end{table*}

\subsection{Ablation Study}

As shown in Table~\ref{tab:ablation_drl} and Table~\ref{tab:ablation}, to verify the effectiveness of the DRL module and the sequential inference strategy, several additional models are trained for comparison.

\noindent
\textbf{DRL module.}
Accurate degradation representations and appropriate training strategies can effectively guide the QECV network to reduce compression artifacts. As shown in Table~\ref{tab:ablation_drl}, taking the DRL module pre-trained only with classification learning as the baseline, introducing contrastive learning into the DRL training process yields stable performance improvements across most QPs (27–42), with an average increase of 0.03 dB in PSNR.
Furthermore, compared with training DRL from scratch, the proposed training strategy achieves better performance under most QP settings. From QP27 to QP42, the frozen pretrained setting improves the average PSNR gain by 0.05 dB compared with the training-from-scratch setting. This suggests that freezing the pretrained DRL module can provide a more stable degradation prior for the enhancement network.

\noindent
\textbf{STDA block.}
The results in Table~\ref{tab:stda_ablation} verify the effectiveness of introducing the degradation tensor into STDA. Using both the degradation tensor and vector consistently achieves better performance across all QPs, yielding an average gain of 0.04 dB in PSNR and 0.044 in SSIM compared with using only the degradation vector. These improvements indicate that the global degradation vector mainly captures frame-level compression severity, while the degradation tensor further provides spatially varying degradation cues for finer-grained feature modulation. This observation is consistent with our motivation that a single global vector is insufficient to characterize the non-uniform compression artifacts within a frame. By incorporating the degradation tensor, the model can better adapt to locally varying distortions and thus achieve more effective quality enhancement.

\noindent
\textbf{Sequential Inference Strategy.}
Videos compressed at higher QPs exhibit more severe degradations, requiring more artifact reduction stages for effective enhancement. The sequential inference strategy adapts computational complexity based on estimated degradation levels. As shown in Table~\ref{tab:ablation}, the results on the second row indicate that inference time increases with QP for models using sequential inference strategy. For instance, inference time at QP37 is 17\% less than that at QP42 (1.0 h vs. 1.2 h). By employing the sequential inference strategy, the inference time have been reduced across QP22, QP27, QP32, and QP37. For example, inference time is halved (1.2 h to 0.6 h) at QP22. In terms of enhancement performance, slight degradation is observed only under low compression levels after applying this strategy. Specifically, in terms of PSNR, improvements of 0.04 dB and 0.03 dB are achieved at QP32 and QP37, respectively, while a drop of 0.08 dB is observed at QP22. This is because lightly compressed videos already possess relatively high PSNR and SSIM values, leaving limited room for further improvement. These results demonstrate that our model strikes a balance between enhancement performance and inference time, if the sequential inference strategy is removed, the additional PSNR gain is marginal, while the inference time is substantially reduced.

\noindent
\textbf{Adaptive Computation Mechanism.}
To further evaluate the effectiveness of the proposed Sequential Inference Strategy, we compare it with two additional adaptive computation mechanisms on the RaceHorses (416$\times$240) sequence, as shown in Table~\ref{tab:exit_strategy_comparison}. The first one is an IQA-based strategy inspired by RBQE, where the intermediate enhanced result after each artifact reduction stage is evaluated by an IQA module, and the enhancement process is terminated once the IQA score satisfies a predefined threshold. We test two threshold settings, i.e., $T_Q=0.90$ and $T_Q=0.85$, to analyze their influence on enhancement performance and inference cost. The second one is a learnable-gate strategy, where a lightweight gating branch is inserted after each artifact reduction stage to predict whether the current inference process should continue or stop. During inference, the stopping stage is determined by the gate output instead of being directly fixed by the predicted degradation level.

The results show that the proposed Sequential Inference Strategy achieves a better quality-efficiency trade-off than the compared adaptive computation mechanisms. Compared with the IQA-based strategy, our method obtains higher PSNR gains with significantly lower inference time. For example, at QP22, the IQA-based strategy with $T_Q=0.90$ achieves a PSNR gain of 0.64 dB and requires 69.28 min, while the proposed Sequential Inference Strategy achieves 0.81 dB with only 5.95 s. At QP42, the same IQA-based strategy obtains a PSNR gain of 0.93 dB and requires 71.24 min, whereas our method achieves 1.04 dB with only 11.34 s. This is because the IQA-based strategy needs to repeatedly evaluate intermediate enhanced results, which introduces substantial computational overhead. Moreover, changing the threshold can reduce the inference time to some extent, but the performance remains lower than that of our method, indicating that the IQA-based decision is sensitive to manually predefined thresholds and is not directly optimized for the final restoration metrics. Compared with the learnable-gate strategy, the proposed Sequential Inference Strategy achieves almost the same enhancement performance but consistently requires less inference time. For instance, at QP27, both methods achieve a PSNR gain of 0.94 dB, while our method reduces the inference time from 11.04 s to 6.86 s. At QP42, both methods achieve a PSNR gain of 1.04 dB, while our method reduces the inference time from 20.39 s to 11.34 s. These results demonstrate that the proposed Sequential Inference Strategy provides more stable and efficient inference control with lower decision-making cost.

\section{Conclusion}
In this paper, we propose a Fine-Grained Degradation-Guided Sequential Inference framework for blind quality enhancement of compressed video. Our method introduces a Degradation Representation Learning (DRL) module that leverages both contrastive and classification losses to extract fine-grained, multi-scale degradation representations, effectively guiding the enhancement network to adapt to diverse artifact characteristics. To improve computational efficiency, a sequential inference strategy dynamically adjusts processing stages according to the detected degradation severity. Furthermore, we design a dual-branch artifact reduction structure that integrates global contextual information with local spatial details, enabling comprehensive exploitation of spatiotemporal dependencies. Extensive experiments on the MFQE 2.0 dataset and Vimeo-90k dataset validate that our approach achieves state-of-the-art performance in blind QECV tasks.

\bibliographystyle{IEEEtran}
\bibliography{main}

\begin{IEEEbiography}[{\includegraphics[width=1in,height=1.25in,clip,keepaspectratio]{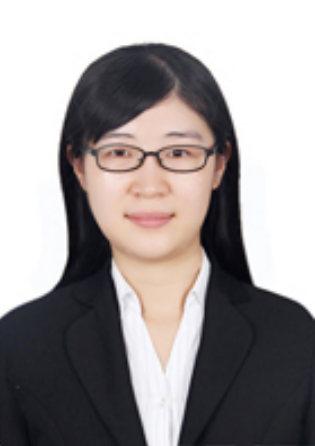}}]{Li Yu}
(Member, IEEE) received the B.S. degree from Soochow University, Suzhou, China, in 2012, and the Ph.D. degree in electrical engineering and electronics from the University of Liverpool, Liverpool, U.K., in 2017. From 2017 to 2018, she was a Postdoctoral Researcher with the Department of Signal Processing, Tampere University of Technology, Tampere, Finland. Since 2018, she has been a Faculty Member with Nanjing University of Information Science and Technology, Nanjing, China. Her research interests include video streaming, video coding, compressed video quality enhancement, computer vision, and deep learning.
\end{IEEEbiography}

\begin{IEEEbiography}[{\includegraphics[width=1in,height=1.25in,clip,keepaspectratio]{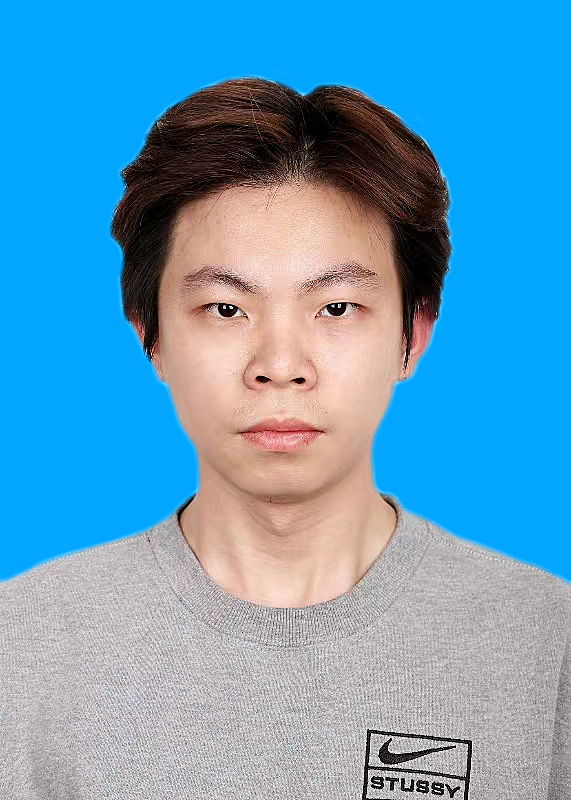}}]{Yingbo Zhao}
is currently pursuing a master’s
degree in Computer Science at university of Nanjing University of Information Science and Technology(NUIST), China. His research interests include compressed video quality enhancement and machine learning.
\end{IEEEbiography}

\begin{IEEEbiography}[{\includegraphics[width=1in,height=1.25in,clip,keepaspectratio]{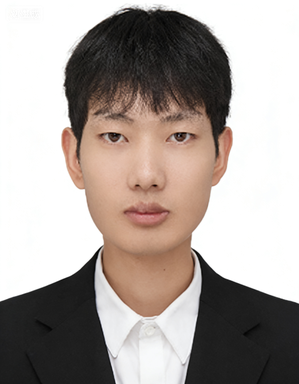}}]{Shiyu Wu}
received the master’s degree from the School of Software, Nanjing University of Information Science and Technology, Nanjing, China. His research interests include video restoration, semi-supervised learning, and machine learning.
\end{IEEEbiography}

\begin{IEEEbiography}[{\includegraphics[width=1in,height=1.25in,clip,keepaspectratio]{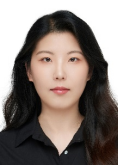}}]{Siyue Yu}
(Member, IEEE) recieved B.S. degree from Xi’an Jiaotong-Liverpool University (Department of Electrical and Electronic Engineering). In December 2022, she received a PhD degree from the University of Liverpool (School of Electrical and Electronic Engineering). Joined Xi’an Jiaotong-Liverpool University in January 2023 as an assistant professor. Her current research interests include computer vision, including weakly supervised semantic segmentation, salient object detection, and anomaly detection.
\end{IEEEbiography}

\begin{IEEEbiography}[{\includegraphics[width=1in,height=1.25in,clip,keepaspectratio]{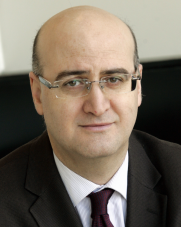}}]{Moncef Gabbouj}
(Fellow, IEEE) is Professor of Information Technology at the Department of Computing Sciences, Tampere University, Finland. He was an Academy of Finland Professor. His research interests include Big Data analytics, multimedia analysis, artificial intelligence, machine learning, pattern recognition, nonlinear signal processing, video processing, and coding. Dr. Gabbouj is a Fellow of the IEEE and Asia-Pacific Artificial Intelligence Association. He is member of the Academia Europaea, the Finnish Academy of Science and Letters and the Finnish Academy of Engineering Sciences. Dr. Gabbouj is the Finland Site Director of the NSF IUCRC funded Center for Big Learning.
\end{IEEEbiography}

\begin{IEEEbiography}[{\includegraphics[width=1in,height=1.25in,clip,keepaspectratio]{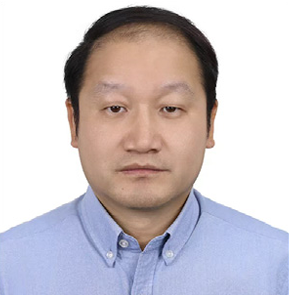}}]{Qingshan Liu}
(Senior Member, IEEE) received the Ph.D. degree in pattern recognition and intelligence systems from Chinese Academy of Sciences, Beijing, China, in 2003. After the doctoral studies, he joined the National Laboratory of Pattern Recognition, Chinese Academy of Sciences. From 2004 to 2005, he was an Associate Researcher with the Multimedia Laboratory, The Chinese University of Hong Kong, Hong Kong. From 2006 to 2011, he worked with the Department of Computer Science, Computational Biomedicine Imaging and Modeling Center, Rutgers University, New Brunswick, NJ, USA, and The State University of New Jersey, Piscataway, NJ, USA. From 2011 to 2023, he was a Professor with the School of Computer and Software, Nanjing University of Information Science and Technology, Nanjing. He is currently a Professor with the School of Computer Science, Nanjing University of Posts and Telecommunications. His research interests include pattern recognition and image and video understanding.
\end{IEEEbiography}

\vfill

\end{document}